\PassOptionsToPackage{svgnames,table}{xcolor}
\documentclass[11pt,letterpaper]{mystyle}

\usepackage[comma,authoryear,compress]{natbib}
\usepackage[all]{hypcap}
\usepackage{placeins}
\usepackage{float}

\usepackage{amsmath,amsfonts,bm}

\def\eqref#1{equation~\ref{#1}}

\def\1{\bm{1}}

\DeclareMathAlphabet{\mathsfit}{\encodingdefault}{\sfdefault}{m}{sl}
\SetMathAlphabet{\mathsfit}{bold}{\encodingdefault}{\sfdefault}{bx}{n}

\hypersetup{
  colorlinks=true,
  citecolor=SEUGreenDark,
  linkcolor=SEUGreenDark,
  urlcolor=SEUGreen
}
\definecolor{bestcell}{HTML}{F2F2FE}
\definecolor{mypurple}{HTML}{E6F2EC}

\title{\textbf{SelectTTRL: Representation-Guided Data Selection for Efficient Test-Time Reinforcement Learning}}
\runningtitle{SelectTTRL: Representation-Guided Data Selection for Efficient Test-Time Reinforcement Learning}
\author{Yi Li$^{1}$, Linan Yue$^{1}$, and Weibo Gao$^{2}$}
\affil[1]{Southeast University}
\affil[2]{The Hong Kong Polytechnic University}
\correspondingauthor{Linan Yue, Email: \href{mailto:lnyue@seu.edu.cn}{lnyue@seu.edu.cn}}

\begin{document}
\begin{abstract}
Test-Time Reinforcement Learning (TTRL) enables large language models to adapt directly on unlabeled test problems by constructing pseudo-labels from multiple sampled responses. However, existing TTRL methods typically apply the same expensive sampling and optimization procedure to all test problems, despite substantial differences in their training utility. In this paper, we study whether such problems can be identified before costly test-time training, without access to ground-truth answers. We uncover a stable relationship between internal response representations and problem-level empirical accuracy. Specifically, we define COS as the cosine similarity between a model's response representation and a correctness-related direction constructed from an independent question-answering dataset. COS consistently correlates with problem solvability across different models, datasets, and experimental settings, and we further provide a theoretical analysis explaining this relationship. Building on this finding, we propose SelectTTRL, a representation-guided data selection method that ranks test problems using COS and applies TTRL only to a selected percentile interval. SelectTTRL leaves the standard TTRL pipeline unchanged for the selected problems, while avoiding unnecessary multi-response sampling and reinforcement learning updates on the remaining ones. Experiments across multiple reasoning benchmarks demonstrate that SelectTTRL substantially reduces test-time computation while maintaining or improving the performance of standard TTRL. Code is available at \url{https://github.com/heyiliyi/SelectTTRL-public}.
\end{abstract}
\maketitle

\section{Introduction}

Large language models (LLMs) have demonstrated strong capabilities on complex reasoning tasks~\citep{qwen2024qwen25,yang2025qwen3}, yet further improving their reasoning ability often relies on additional supervision signals~\citep{shao2024deepseekmath,guo2025deepseekr1}. Test-Time Reinforcement Learning (TTRL)~\citep{zuo2025ttrl} provides a new paradigm for adapting models directly on unlabeled test problems. Specifically, TTRL samples multiple responses for each test problem, derives pseudo-labels through majority voting, and constructs reward signals for reinforcement learning without access to the ground-truth answers. In this way, TTRL enables models to further improve their reasoning ability by exploiting the test data themselves~\citep{hosseini2026cgttrl,yan2026scrl}.

However, TTRL typically incurs substantial test-time computational overhead~\citep{liu2025ettrl}. 
Specifically, to obtain reliable training signals, each problem requires multiple inference samples followed by reinforcement learning updates. When the test set is large, repeatedly sampling and training on every problem introduces substantial additional cost. More importantly, existing TTRL methods typically allocate similar computation across problems, overlooking differences in their value for test-time training. Not every problem is equally worth optimizing. For problems the model can already solve reliably, additional sampling and updates may provide only marginal gains. In contrast, for difficult problems, repeated sampling may produce highly consistent yet incorrect answers, causing erroneous pseudo-labels to be further reinforced during training~\citep{yu2026ddrl,lin2026ttrlguard}. Therefore, uniformly applying the full TTRL procedure to all test problems may waste computation and allocate the limited test-time budget to problems with low training utility. This motivates a natural question:


\emph{Can we identify which test problems are worth training on before performing expensive multi-response sampling and reinforcement learning, without access to ground-truth answers?} 

To answer this question, we investigate whether internal model representations can provide a prior signal of test-problem solvability~\citep{zou2023representation,azaria2023internal}. Specifically, we construct a correctness-related direction from an independent question-answering dataset~\citep{panickssery2024caa}. For each test problem, we then compute the cosine similarity between this direction and the model's internal response representation. We denote this similarity as \textbf{COS}.
We observe a stable relationship between COS and problem-level empirical accuracy. Problems with higher COS generally achieve higher overall accuracy, while those with lower COS are solved less reliably. This relationship holds consistently across different models, datasets, and experimental settings. We further provide a theoretical analysis to explain why this correlation emerges. Our analysis suggests that internal representations contain information related to the relative solvability of test problems. Therefore, COS provides a representation-level signal for estimating and ranking test problems without access to their ground-truth answers.

Building on this finding, we propose \textbf{SelectTTRL}, a representation-guided data selection method for efficient test-time reinforcement learning. Before expensive multi-response sampling, SelectTTRL ranks test problems according to COS. It then selects only problems within a prescribed percentile interval for subsequent training. For the selected problems, the standard TTRL pipeline remains unchanged, including multi-response sampling, pseudo-label construction, and policy optimization. In this way, SelectTTRL avoids performing costly sampling and reinforcement learning updates over the entire test set. 

Our main contributions are summarized as follows:
\begin{itemize}[leftmargin=*]
\item In this paper, we uncover a stable relationship between internal response representations and problem-level empirical accuracy, and formalize it as \textbf{COS}, a representation-level signal for test-problem solvability. Besides, we provide a theoretical analysis explaining why COS reflects the relative solvability of test problems.

\item Building on COS, we propose SelectTTRL. It ranks and selects test problems before the expensive multi-response sampling stage of standard TTRL, and performs subsequent reinforcement learning only on the selected subset. This reduces test-time training cost without modifying the underlying TTRL optimization pipeline.

\item Experiments across multiple reasoning benchmarks show that SelectTTRL substantially reduces test-time computation while maintaining or even improving the performance of standard TTRL.

\end{itemize}

\section{Related Work}

\textbf{Test-time reinforcement learning.}
Test-time reinforcement learning (TTRL) adapts language models to unlabeled test problems by constructing pseudo-supervision from their own sampled responses, typically through majority-voted answers~\citep{zuo2025ttrl}. Existing work mainly improves TTRL from two perspectives. The first focuses on improving the learning signal within each problem. Entropy-guided sampling and contextual guidance regulate exploration during rollout generation~\citep{liu2025ettrl,hosseini2026cgttrl}, while selective labeling and confidence- or consistency-based consensus aim to mitigate noisy pseudo-supervision~\citep{yan2026scrl,wang2026scope,yu2026ddrl,wang2026selfharmony}. These methods improve how sampled responses are generated or converted into training signals, rather than selecting a fixed subset of test problems before adaptation.
A second line of work adaptively controls whether or how a problem contributes to test-time learning. TTRL-Guard adjusts or suspends updates according to consensus dynamics~\citep{lin2026ttrlguard}, while DiSCTT routes problems to pseudo-label SFT or RL based on response agreement~\citep{moradi2026disctt}. However, such decisions are made only after multiple responses have already been sampled, meaning that much of the test-time computation has already been incurred. Moreover, response agreement does not necessarily indicate that a problem is informative for subsequent optimization. In contrast, we consider \emph{pre-selection}: before  multi-response sampling and policy updates, we rank the full set of test problems using representations obtained from an initial response, and apply standard TTRL only to the selected subset. This separates the question of \emph{which problems to optimize on} from modifications to the TTRL learning rule itself.

\textbf{Data selection for LLM training.}
Data selection reduces post-training cost by identifying compact subsets that preserve or improve learning effectiveness~\citep{albalak2024selection}. A broad family of methods scores examples according to properties such as quality, difficulty, uncertainty, and diversity~\citep{liu2024deita,li2024ifd,zhao2025uforl}. Other approaches estimate sample utility through gradients or influence functions. For example, LESS selects instruction-tuning examples according to their gradient influence on target tasks, but requires warm-up training and per-example gradient computation~\citep{xia2024less}. Related methods similarly exploit gradient alignment or influence estimates to identify useful training subsets~\citep{zhang2025tagcos,li2026learnalign}.
More recent work explores selection signals that can be obtained without additional trial training. AIR estimates example utility through loss changes induced by retrieval-head masking~\citep{liu2025air}, while SHIFT combines hidden-state dynamics from a single rollout with diversity-aware selection~\citep{wu2026shift}. These methods reduce the overhead of estimating data utility. However, their downstream optimization still relies on externally available supervision, such as supplied reasoning traces or verifiable rewards. Our setting is different: in TTRL, neither selection nor subsequent learning has access to ground-truth correctness. We therefore study whether an internal representation signal can identify useful test problems \emph{before} pseudo-label construction, under fully unlabeled test-time learning.

\textbf{Representation engineering.}
Representation engineering studies how semantic or behavioral properties are encoded in model activations and how such representations can be read or manipulated~\citep{zou2023representation}. Prior work has shown that hidden states contain predictive signals associated with properties such as truthfulness and correctness~\citep{azaria2023internal}, while contrastive activation addition constructs semantic directions for controlling model behavior~\citep{panickssery2024caa}. Related activation-based methods intervene on internal representations to improve reasoning or elicit different reasoning behaviors~\citep{hojer2025reasoning,tang2025glore,li2025featuresteering}. Sparse-autoencoder-based approaches further identify and manipulate interpretable features associated with knowledge use, semantic attributes, and reasoning strategies~\citep{zhao2025spare,he2025saessv,fang2026saesteering}.
Our use of representations is complementary to these intervention-oriented approaches. Rather than modifying activations during generation, we use a separately calibrated correctness-related direction as a semantic reference for measuring the representation of an initial response. The resulting cosine-similarity signal, COS, is then used to rank unlabeled test problems before TTRL begins. This extends representation-based analysis from predicting or steering model behavior to allocating test-time optimization effort across problems.

\section{Framework of SelectTTRL}
\label{sec:SelectTTRL-method}
Given an unlabeled target set $\mathcal Q=\{q_i\}_{i=1}^{M}$ and an initial
policy $\pi_{\theta_0}$, our goal is to identify which target questions are
most suitable for test-time reinforcement learning before expensive
multi-rollout sampling and policy optimization are performed.
The key challenge is that ground-truth correctness is unavailable on the
target set. We therefore seek a lightweight internal signal that can be
computed from an initial model response and used to estimate the relative
training value of different questions.

SelectTTRL addresses this problem with a representation-based score,
which we call \textbf{COS}. COS measures how strongly the representation of
a target response aligns with a separately calibrated
correctness-related direction. We first construct this direction from a
small labeled calibration set and then keep it fixed throughout target-set
selection. For each target question, we generate only an initial response,
compute its COS score, and rank questions accordingly. We retain a
prescribed percentile interval and apply standard TTRL
reward construction and policy optimization only to the selected subset.
Because selection is performed before TTRL rollout collection, excluded
questions incur neither multi-response sampling nor policy updates.


\subsection{Extracting a correctness-related direction}
\label{sec:SelectTTRL-direction}
To obtain a semantic reference for correctness, we construct a
correctness-related direction using the mean-difference procedure of
CAA~\citep{panickssery2024caa}. Let
$\mathcal D_{\mathrm{cal}}={(x_i,y_i^+,y_i^-)}_{i=1}^{n}$ denote a
separate labeled calibration set derived from
TruthfulQA~\citep{lin2022truthfulqa}, where $y_i^+$ and $y_i^-$ are a
correct and an incorrect answer to question $x_i$, respectively.
For each calibration example, we place the two candidate answers in
randomized option positions within the same prompt. Keeping the prompt and
option order fixed, we separately teacher-force the answer letter
corresponding to the correct and incorrect candidates. We then extract the
layer-$\ell$ residual-stream activations at the answer-letter position,
denoted by $\mathbf h_{\ell,i}^{+}$ and $\mathbf h_{\ell,i}^{-}$.
The correctness-related direction is defined as the normalized mean
difference between these activations:
\begin{equation}
\mathbf d_\ell=\frac{1}{n}\sum_{i=1}^{n}
\bigl(\mathbf h_{\ell,i}^{+}-\mathbf h_{\ell,i}^{-}\bigr),
\qquad
\mathbf v_\ell=\frac{\mathbf d_\ell}{|\mathbf d_\ell|_2}.
\label{eq:SelectTTRL-direction}
\end{equation}

Intuitively, $\mathbf v_\ell$ captures a representation axis along which
the model internally distinguishes more correct-like from less
correct-like answers. Prior work has shown that language models encode
confidence- and correctness-related information in their internal
representations during generation~\citep{kumaran2026verbalconfidence}.
This motivates using $\mathbf v_\ell$ as a fixed reference direction for
reading correctness-related evidence from responses to unseen target
questions.

Importantly, the labeled calibration set is used only once to estimate
$\mathbf v_\ell$. The direction is then frozen and no target labels are
used during question selection or subsequent TTRL adaptation. In the next
section, we use this direction to define COS for each target response and
rank the target questions before rollout-intensive TTRL begins.

\subsection{Question-level COS scoring}
\label{sec:SelectTTRL-scoring}
For a response $y$ generated by $\pi_{\theta_0}$ for question $q$, let
$\mathbf h_\ell(q,y)$ denote the layer-$\ell$ residual-stream activation
at the fixed answer position. Its COS score is
\begin{equation}
  X(q,y)=\frac{\mathbf h_\ell(q,y)^\top\mathbf v_\ell}
  {\|\mathbf h_\ell(q,y)\|_2\,\|\mathbf v_\ell\|_2}.
  \label{eq:SelectTTRL-cos}
\end{equation}
We obtain a question-level score either from a single greedy response
$y_q^{\mathrm{g}}$ or by averaging COS over $K$ sampled responses under a
fixed decoding protocol:
\begin{equation}
  c_{\mathrm{g}}(q)=X(q,y_q^{\mathrm{g}}),\qquad
  c_K(q)=\frac{1}{K}\sum_{j=1}^{K}X(q,y_{qj}).
  \label{eq:SelectTTRL-score-variants}
\end{equation}
Both scoring variants are positively correlated with question-level
accuracy in our experiments.

\subsection{Selecting questions for TTRL}
\label{sec:SelectTTRL-selection}

COS provides an ordering of target questions according to the model's
internal correctness-related signal. However, our goal is not simply to
select questions with the highest or lowest COS. Questions that the model
already solves reliably may provide limited room for improvement, whereas
questions at the opposite extreme are more likely to yield unreliable
pseudo-labels during TTRL. We therefore focus the optimization budget on an
intermediate region of the COS ranking, where sufficient learning
opportunity and pseudo-label reliability are more likely to coexist.

Let $q_{(1)},\ldots,q_{(M)}$ denote the target questions sorted in ascending
order of $c(q)$, with ties broken by a fixed question identifier. For prescribed
percentile endpoints $0\leq a<b\leq1$, we select
\begin{equation}
  \mathcal S_{a,b}
  =\bigl\{q_{(i)}:\lfloor aM\rfloor<i\leq\lfloor bM\rfloor\bigr\},
  \qquad
  \rho=\frac{|\mathcal S_{a,b}|}{M}.
  \label{eq:SelectTTRL-selection}
\end{equation}
The percentile interval is fixed before target adaptation and is selected
without access to target ground-truth answers. Only questions in
$\mathcal S_{a,b}$ undergo subsequent TTRL rollout collection and policy
updates, whereas all questions in $\mathcal Q$ remain in the evaluation
set. Thus, excluded questions incur only the cost of the initial response
used for COS scoring.

\subsection{TTRL on the selected subset}
\label{sec:SelectTTRL-update}

After obtaining the selected subset \(\mathcal S_{a,b}\), we apply the standard TTRL procedure \citep{zuo2025ttrl} without modifying its reward construction or policy optimization.
For each selected question \(q\), the rollout policy \(\pi_{\theta_{\mathrm{old}}}\) generates a vote pool \(\{y_{q,j}\}_{j=1}^{N_{\mathrm{vote}}}\). Let \(f(\cdot)\) denote the function that extracts and normalizes the final answer. The majority-voted answer is used as the pseudo-label, and each rollout receives a binary reward according to its agreement with this pseudo-label:
\begin{equation}
  \hat a_q=\operatorname*{arg\,max}_{a}\sum_{j=1}^{N_{\mathrm{vote}}}
  \mathbf{1}\{f(y_{q,j})=a\},
  \qquad
  r_q(y)=\mathbf{1}\{f(y)=\hat a_q\}.
  \label{eq:SelectTTRL-reward}
\end{equation}
Following the standard vote-then-sample protocol, the pseudo-label is computed from the full vote pool, after which \(G\leq N_{\mathrm{vote}}\) responses are sampled for optimization. Their rewards are converted into group-relative advantages using GRPO \citep{shao2024deepseekmath}, where $\hat A_{q,i}=\frac{r_{q,i}-\bar r_q}{s_q+\epsilon_{\mathrm{num}}}, \bar r_q=\frac{1}{G}\sum_{i=1}^{G}r_{q,i}$, and \(s_q\) is the reward standard deviation within the update group.
Let \(\ell_{\mathrm{TTRL}}(q;\theta)\) denote the standard TTRL policy loss for question \(q\). SelectTTRL optimizes this loss only over the selected subset:
\begin{equation}
  \mathcal L_{\mathrm{SelectTTRL}}(\theta)
  =\mathbb E_{q\sim\operatorname{Unif}(\mathcal S_{a,b})}
  [\,\ell_{\mathrm{TTRL}}(q;\theta)\,].
  \label{eq:SelectTTRL-objective}
\end{equation}
Thus, SelectTTRL leaves the underlying TTRL learning rule unchanged and only determines which target questions receive rollout and update budgets.

\section{Theoretical Analysis of COS}

\label{sec:SelectTTRL-theory}
Our goal is to explain why COS can transfer across tasks. 
We assume that model representations share a correctness-related component across tasks, which COS reads through a contrastively extracted direction. This direction can capture multiple latent features, further analyzed in RQ3. The model, layer, and readout protocol are fixed. Let
$U=\mathbf h_\ell/\|\mathbf h_\ell\|_2$, $v=\mathbf v_\ell$, and $X=v^\top U$.

\newtheorem{cosassumption}{Assumption}[section]
\newtheorem{coslemma}[cosassumption]{Lemma}
\newtheorem{coscondition}[cosassumption]{Condition}
\newtheorem{costheorem}[cosassumption]{Theorem}

\subsection{Shared evidence geometry}

\begin{cosassumption}[Spherical evidence model]
\label{ass:cos-evidence}
For each target question $q$ in domain $D$, the normalized response
representation follows $U\mid q\sim p_D(\cdot\mid t_q)$, where
\begin{equation}
  p_D(U\mid t)=b_D(U)\exp\{t\,u^\top U-A_D(tu)\},
  \qquad \|u\|_2=1.
  \label{eq:cos-evidence-model}
\end{equation}
Here $u$ is a shared evidence direction, $t_q$ is a question's evidence
state, $b_D$ is a domain-specific background density on the unit sphere,
and $A_D$ is the log-normalizer.
\end{cosassumption}

Increasing $t$ favors representations with larger evidence projection
$T=u^\top U$, while retaining the background geometry. The uniform-background
case reduces to a von Mises--Fisher distribution; a general $b_D$ permits
anisotropy. This model concerns evidence states rather than two classes
with equal concentration.
Let $m_D(t)=\mathbb E_t[U]$ and
$\Sigma_D(t)=\operatorname{Cov}_t(U)$. The exponential-family identities
give~\citep{wainwright2008graphical}
\begin{equation}
  \frac{d m_D(t)}{dt}=\Sigma_D(t)u,
  \qquad
  \frac{d\,\mathbb E_t[X]}{dt}=v^\top\Sigma_D(t)u.
  \label{eq:cos-response}
\end{equation}
Thus, the response of COS depends on the extracted direction and
the target representation geometry. The key transfer question is when
source extraction yields $v^\top\Sigma_D(t)u>0$ on the target domain.

\subsection{Contrastive extraction and positive transfer}

Eq.~(\ref{eq:SelectTTRL-direction}) maximizes the average signed projection of source contrasts. Under the controlled source construction in Appendix~\ref{app:cos-source}, the population raw-activation contrast is:
\begin{equation}
  d_S^\star=M_Su+e_S,
  \qquad
  M_S=\bar R_S\int_{t_-}^{t_+}\Sigma_S(r)\,dr\succeq0,
  \label{eq:cos-source-response}
\end{equation}
where $t_+>t_-$, $\bar R_S>0$, and $e_S$ captures radial and unmatched source effects. Let $\hat d_S$ be the empirical contrast and
$v=\hat d_S/\|\hat d_S\|_2$. We consider three hypotheses for how this direction reads shared evidence.

\textbf{$\mathsf H_1$: Uniform source gain.}
Source accuracy $a_S$ changes only the overall contrast strength:
\begin{equation}
  \hat d_S=\gamma(a_S)(d_0+\xi_S),\qquad
  \gamma(a_S)>0,\quad \xi_S\perp\!\!\!\perp a_S.
  \label{eq:cos-uniform-gain}
\end{equation}
Here $d_0$ is fixed. Normalization removes $\gamma$, predicting source-accuracy-independent
readout quality on fixed target representations. The overall positive
trend in Table~\ref{tab:cos-source-comparison} argues against this prediction.

\textbf{$\mathsf H_2$: Single-factor linear response.}
A single factor linearly mediates changes in both COS and correctness.
Across feature interventions $j$, this predicts
\begin{equation}
  \Delta P_j=b\,\Delta C_j,\qquad b>0,
  \label{eq:cos-single-factor}
\end{equation}
where $C$ and $P$ denote mean COS and accuracy.
Table~\ref{tab:sae-feature-ablation} does not follow this proportionality:
ablations 31368 and 48965 produce the same accuracy decrease but markedly
different COS changes.

\textbf{$\mathsf H_3$: Multicomponent positive compatibility.}
Source reliability affects the relative strength of shared evidence and
source-specific residuals. Multiple components may contribute to the readout;
positive transfer requires their combined response to remain positive.
We formalize this hypothesis for the actual extracted contrast.

\begin{coscondition}[Positive compatibility]
\label{cond:cos-compatibility}
Write $\hat d_S=\alpha_Su+r_S$, where $r_S\perp u$ and $\alpha_S>0$.
On an interval containing the question states, let
$\beta_D(t)=u^\top\Sigma_D(t)u>0$. For some $\delta\in(0,1]$, assume
\begin{equation}
  r_S^\top\Sigma_D(t)u
  \geq -(1-\delta)\alpha_S\beta_D(t).
  \label{eq:cos-compatibility}
\end{equation}
\end{coscondition}

This condition allows rotated directions and nonzero residuals: their
adverse response must not cancel the shared-evidence contribution.
Tables~\ref{tab:sae-feature-ablation} and~\ref{tab:cos-source-comparison}
motivate this multicomponent account. Stronger evidence can improve
directional stability by reducing relative extraction error
(Appendix~\ref{app:cos-hypotheses}).

\subsection{From evidence to question accuracy}

\begin{coslemma}[Positive response margin]
\label{lem:cos-positive-response}
Under Assumption~\ref{ass:cos-evidence} and
Condition~\ref{cond:cos-compatibility},
\begin{equation}
  F_D'(t)=v^\top\Sigma_D(t)u
  \geq \frac{\delta\alpha_S\beta_D(t)}{\|\hat d_S\|_2}>0,
  \qquad F_D(t)=\mathbb E_t[X].
  \label{eq:cos-positive-margin}
\end{equation}
\end{coslemma}

\begin{coscondition}[Reliability link]
\label{cond:cos-transfer}
For $Y\in\{0,1\}$ denoting objective correctness, assume
$P(Y=1\mid T,Q=q)=g_D(T)$ for each question, with $g_D$ strictly increasing on the support
of $T$.
\end{coscondition}

Define $G_D(t)=\mathbb E_t[g_D(T)]$. The response identity yields
\begin{equation}
  G_D'(t)=\operatorname{Cov}_t(g_D(T),T)>0.
  \label{eq:cos-evidence-correctness}
\end{equation}
Thus both expected COS and objective accuracy increase with evidence.

\begin{costheorem}[Question-level correctness transfer]
\label{thm:cos-transfer}
Under Assumption~\ref{ass:cos-evidence} and
Conditions~\ref{cond:cos-compatibility} and~\ref{cond:cos-transfer}, let the question state $t_Q$ be
non-degenerate. For $K$ responses sampled under the scoring protocol,
and an additional response with correctness $Y'$ independent of them
conditional on~$Q$,
\begin{equation}
  \operatorname{Cov}(c_K(Q),Y')
  =\operatorname{Cov}_Q\bigl(F_D(t_Q),G_D(t_Q)\bigr)>0,\,\,\,\,\,\,
  I(c_K(Q);Y')>0.
  \label{eq:SelectTTRL-positive-association}
\end{equation}
The same positive covariance holds with $Y'$ replaced by
$p_Q=P(Y'=1\mid Q)$.
\end{costheorem}

Proofs are in Appendix~\ref{app:cos-theory}.

\textit{Remark.}
Under these conditions, average COS predicts an independent answer's
correctness without identifying every incorrect response. RQ2 complements
this readout analysis by examining how steering direction and strength
affect reasoning and answer completion. Different domains
need not share a COS threshold. Averaging reduces sampling variation,
while departures from the ideal model can produce local ranking reversals
(Appendix~\ref{app:cos-robustness}).

\section{Experiments}

We evaluate SelectTTRL and analyze the COS signal through four research questions:
\textbf{RQ1:} How does SelectTTRL affect model performance and training efficiency in test-time reinforcement learning?
\textbf{RQ2:} Can interventions along the extracted direction influence the correctness of model responses?
\textbf{RQ3:} To what extent is COS explained by known response representations, and is it simply a combination of interpretable features?
\textbf{RQ4:} What types of source datasets yield directions for which COS positively correlates with question-level accuracy?

\subsection{Experimental Setup}
\textbf{Benchmarks.}    We evaluate SelectTTRL on six reasoning benchmarks:
GPQA-Diamond~\citep{rein2024gpqa}, a challenging and high-quality subset of the
Graduate-Level Google-Proof Question Answering benchmark;
MMLU-Pro STEM, the STEM subset of MMLU-Pro~\citep{wang2024mmlupro};
and four mathematical reasoning benchmarks, AIME 2024 (30 questions)~\citep{aimo2024aime},
AIME 2025~\citep{zhang2025aime}, AMC (83 questions from 2022--2023)~\citep{aimo2024amc},
and MATH-500~\citep{hendrycks2021math}.

\textbf{Baselines.} We use Qwen3-8B~\citep{yang2025qwen3} and Qwen2.5-7B-Instruct~\citep{qwen2024qwen25} as backbones and compare SelectTTRL with three baselines: (1) \textbf{Matched-TTRL}~\citep{zuo2025ttrl}, which applies TTRL to the full unlabeled test set; (2) \textbf{A-PPL}, which selects questions with the highest perplexity of model-generated answers; and (3) \textbf{Random}, which uniformly samples questions without replacement. A-PPL and Random retain 35\% of the question pool, matching SelectTTRL. For all subset-selection methods, only the selection rule varies, while the TTRL procedure, training budget, and hyperparameters are kept fixed.


\textbf{Hyperparameter Settings.} All experiments use non-thinking mode. For SelectTTRL, we set $(a,b)=(0.25,0.60)$, selecting the 25th--60th percentile interval of the ascending COS ranking. We train with AdamW and cosine learning-rate decay, with a peak learning rate of $5\times10^{-7}$. The maximum context and response lengths are 8,192 and 4,096 tokens, respectively. For each question, we sample 32 responses at temperature $0.6$ for pseudo-label voting and use 16 for policy optimization. Experiments are conducted on 4 NVIDIA A800 80GB GPUs.


\textbf{Evaluation Metrics.} We report greedy-decoding accuracy (Greedy),
sampled pass@1 (Mean1), and majority-vote accuracy (Maj.@16). Among them, 
following DeepSeek-R1~\citep{guo2025deepseekr1}, we evaluate sampled pass@1
using 16 responses per question at temperature $0.6$. Maj.@16 measures the
accuracy of the most frequent answer among the same 16 responses.

\begin{table}[t]
\caption{Comparison of TTRL methods on Qwen3-8B across six reasoning benchmarks.}

\label{tab:main-results}
\centering
\setlength{\tabcolsep}{2.3mm}{
     \resizebox{\linewidth}{!}{
\begin{tabular}{@{}lccc ccc ccc@{}}
\toprule
& \multicolumn{3}{c}{MATH-500}
& \multicolumn{3}{c}{GPQA-Diamond}
& \multicolumn{3}{c}{MMLU-Pro STEM} \\
\cmidrule(lr){2-4}
\cmidrule(lr){5-7}
\cmidrule(lr){8-10}
Method
& Greedy & Mean1 & Maj.@16
& Greedy & Mean1 & Maj.@16
& Greedy & Mean1 & Maj.@16 \\
\midrule

Matched-TTRL
& 86.20 & 87.18 & 92.40
& 51.01 & 51.33 & 57.07
& 75.30 & 76.61 & 79.52 \\

Random
& 86.00 & 86.71 & 91.20
& 49.49 & 50.95 & 54.55
& 75.70 & 75.41 & 79.32 \\

A-PPL
& 87.40 & 86.84 & 91.00
& 53.54 & 50.69 & 56.06
& 75.10 & 76.58 & 81.53 \\

\rowcolor{mypurple}
SelectTTRL
& 86.20 & 86.55 & 92.60
& 51.52 & 50.73 & 58.08
& 74.10 & 75.77 & 80.12 \\

\midrule[0.8pt]

& \multicolumn{3}{c}{AIME 2024}
& \multicolumn{3}{c}{AIME 2025}
& \multicolumn{3}{c}{AMC} \\
\cmidrule(lr){2-4}
\cmidrule(lr){5-7}
\cmidrule(lr){8-10}
Method
& Greedy & Mean1 & Maj.@16
& Greedy & Mean1 & Maj.@16
& Greedy & Mean1 & Maj.@16 \\
\midrule

Matched-TTRL
& 36.67 & 40.42 & 50.00
& 23.33 & 23.54 & 23.33
& 69.88 & 67.85 & 78.31 \\

Random
& 33.33 & 36.04 & 46.67
& 16.67 & 17.71 & 16.67
& 65.06 & 63.86 & 73.49 \\

A-PPL
& 33.33 & 37.29 & 46.67
& 16.67 & 17.50 & 16.67
& 60.24 & 65.74 & 73.49 \\

\rowcolor{mypurple}
SelectTTRL
& 43.33 & 38.75 & 50.00
& 23.33 & 19.79 & 20.00
& 66.27 & 66.04 & 74.70 \\

\bottomrule
\end{tabular}}}

\end{table}

\begin{table}[t]
\caption{Comparison of TTRL methods on Qwen2.5-7B-Instruct across six reasoning benchmarks.}

\label{tab:main-results-qwen25}
\centering
\setlength{\tabcolsep}{2.3mm}{
     \resizebox{\linewidth}{!}{
\begin{tabular}{@{}lccc ccc ccc@{}}
\toprule
& \multicolumn{3}{c}{MATH-500}
& \multicolumn{3}{c}{GPQA-Diamond}
& \multicolumn{3}{c}{MMLU-Pro STEM} \\
\cmidrule(lr){2-4}
\cmidrule(lr){5-7}
\cmidrule(lr){8-10}
Method
& Greedy & Mean1 & Maj.@16
& Greedy & Mean1 & Maj.@16
& Greedy & Mean1 & Maj.@16 \\
\midrule

Matched-TTRL
& 78.80 & 78.34 & 82.00
& 35.35 & 35.04 & 35.86
& 64.66 & 65.31 & 69.28 \\

Random
& 79.80 & 77.68 & 80.80
& 36.67 & 36.81 & 40.40
& 64.46 & 63.45 & 67.27 \\

A-PPL
& 78.40 & 78.04 & 82.40
& 32.83 & 35.86 & 37.88
& 64.66 & 62.79 & 66.47 \\

\rowcolor{mypurple}
SelectTTRL
& 79.60 & 78.61 & 83.00
& 36.87 & 35.61 & 38.89
& 62.65 & 64.22 & 68.47 \\

\midrule[0.8pt]

& \multicolumn{3}{c}{AIME 2024}
& \multicolumn{3}{c}{AIME 2025}
& \multicolumn{3}{c}{AMC} \\
\cmidrule(lr){2-4}
\cmidrule(lr){5-7}
\cmidrule(lr){8-10}
Method
& Greedy & Mean1 & Maj.@16
& Greedy & Mean1 & Maj.@16
& Greedy & Mean1 & Maj.@16 \\
\midrule

Matched-TTRL
& 23.33 & 21.04 & 23.33
& 16.67 & 16.46 & 16.67
& 51.81 & 50.23 & 53.01 \\

Random
& 16.67 & 15.21 & 20.00
& 6.67 & 8.96 & 13.33
& 45.78 & 46.08 & 51.81 \\

A-PPL
& 6.67 & 12.92 & 20.00
& 6.67 & 9.17 & 16.67
& 42.17 & 44.95 & 54.22 \\

\rowcolor{mypurple}
SelectTTRL
& 16.67 & 18.33 & 23.33
& 13.33 & 12.08 & 13.33
& 48.19 & 48.27 & 53.01 \\

\bottomrule
\end{tabular}}}

\end{table}


\subsection{Overall Performance (RQ1)}

SelectTTRL performs TTRL on only 35\% of the target questions using pretrained LLMs such as Qwen3-8B and Qwen2.5-7B-Instruct. Tables~\ref{tab:main-results} and~\ref{tab:main-results-qwen25} compare it with full-data TTRL and alternative selection strategies. Overall, SelectTTRL retains most of the gains of full-data TTRL while using substantially fewer questions for adaptation. On MATH-500 and MMLU-Pro STEM, it performs close to Matched-TTRL despite training on only a subset of the target set.
The benefit of selective training is more pronounced on the harder AIME benchmarks. With Qwen3-8B, SelectTTRL outperforms Random and A-PPL in Greedy accuracy on both AIME 2024 and AIME 2025 while remaining competitive with Matched-TTRL. On AIME 2024, it achieves 43.33\% Greedy accuracy, compared with 36.67\% for Matched-TTRL and 33.33\% for both Random and A-PPL. These results suggest that TTRL performance depends not only on the amount of test-time training data, but also on which questions receive the optimization budget.

We attribute this behavior to the quality of COS-based ranking. Questions with intermediate COS values tend to provide a better trade-off between optimization headroom and pseudo-label reliability: very easy questions offer limited additional learning signal, whereas extremely difficult questions are more prone to unreliable consensus rewards. SelectTTRL therefore concentrates TTRL updates on questions that are more likely to provide useful training signals.
The trend is strongest for Qwen3-8B, where COS exhibits a clearer association with question-level correctness. For Qwen2.5-7B-Instruct, the ranking signal is somewhat less discriminative, and SelectTTRL performs closer to full-data TTRL. Nevertheless, the method remains competitive across both backbone LLMs, suggesting that its effectiveness does not rely on a single model-specific ranking behavior.

Notably, all results use the correctness direction calibrated from TruthfulQA. Although alternative calibration sources can yield stronger COS--accuracy correlations (Section~\ref{sec:cos-source}), the fixed TruthfulQA direction is sufficient to support effective question selection across models and benchmarks.

\FloatBarrier
\subsection{Activation Steering and Answer Correctness (RQ2)}
\label{sec:cos-steering}


To examine whether the direction used by COS is merely correlational or can actively influence model behavior, we perform CAA steering~\citep{panickssery2024caa} with Qwen3-8B on MATH-500 and AIME24. We inject the raw CAA vector at layer 21 during generation with $\alpha\in\{-1,-0.5,+0.5,+1\}$, and recompute COS from the generated responses without intervention.

\begin{figure}[htbp]
    \centering
     
    \includegraphics[width=\linewidth]{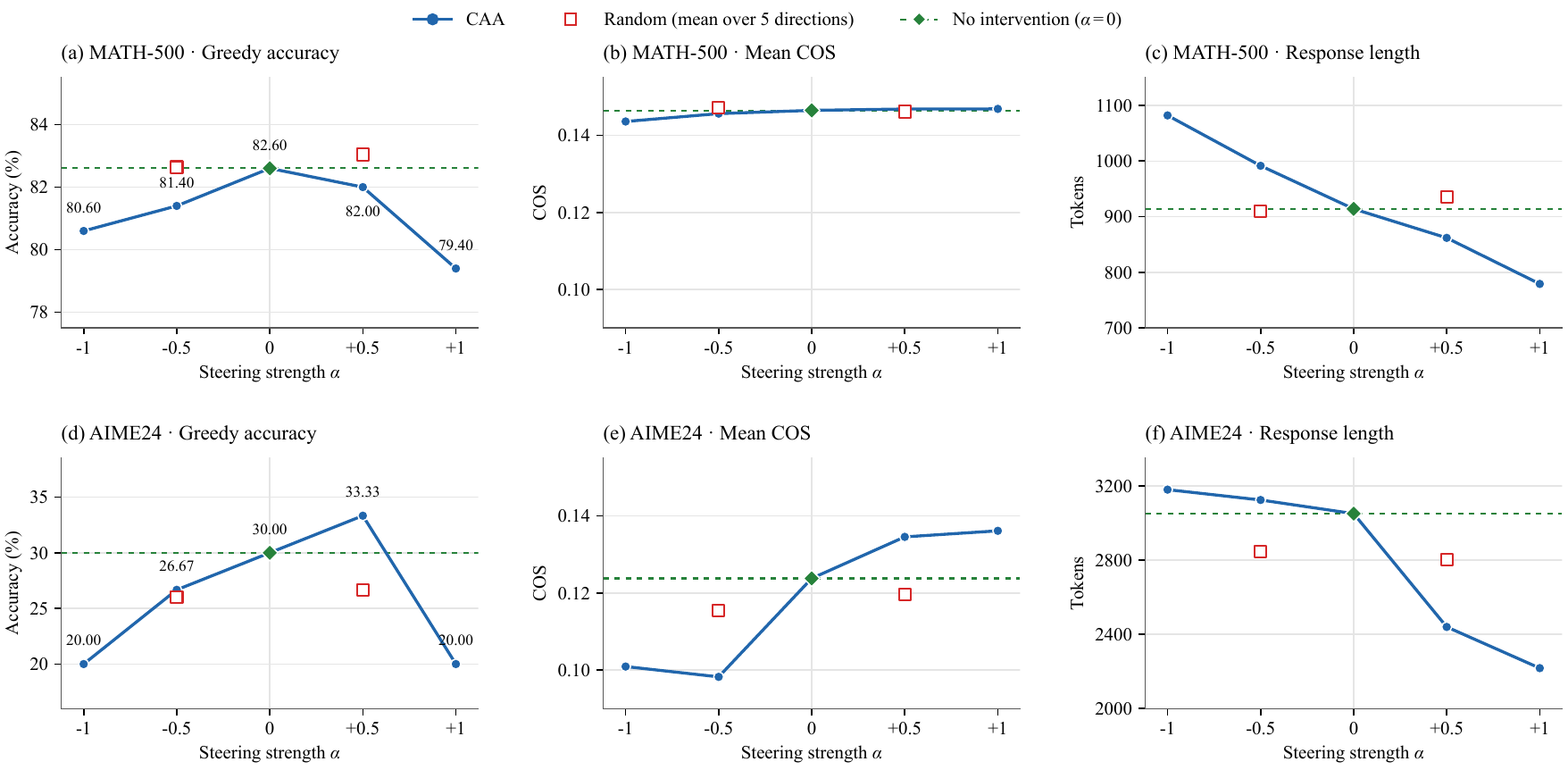}
    
\caption{
CAA and norm-matched random steering results on Qwen3-8B.
Random results are means across five directions.
}
\label{fig:cos-steering-results}
    
\end{figure}

As shown in Figure~\ref{fig:cos-steering-results}, steering along the extracted direction produces systematic changes in both accuracy and reasoning behavior. On AIME24, moderate positive steering ($\alpha=+0.5$) improves accuracy from 30.00\% to 33.33\%, outperforming norm-matched random directions, whereas stronger intervention ($\alpha=+1$) degrades performance. Negative steering consistently lowers COS and tends to lengthen responses, while positive steering increases COS on AIME24 and generally shortens generation. On MATH-500, steering is less beneficial overall, with larger intervention magnitudes causing clearer accuracy degradation.

Response inspection shows that steering affects both calculations and answer completion. For example, on MATH-500, $\alpha=+0.5$ turns a repetitive, truncated response into a correct 578-token solution, whereas $\alpha=-0.5$ causes a previously correct solution to repeat an unfinished step until truncation. Steering can also alter intermediate calculations, such as selecting an incorrect triangle vertex. These cases suggest that the COS direction is behaviorally relevant to correctness, although its effect depends on intervention strength and task.


\FloatBarrier
\subsection{Interpreting COS through Sparse Features (RQ3)}
\label{sec:cos-sae}
To better understand what COS captures internally, we analyze its sparse-feature composition and test whether high-contribution features are behaviorally relevant to answer correctness. Following \citet{marks2025sparsecircuits}, we decompose the residual-stream activation as: 
\begin{equation}
  \mathbf h
  =\underbrace{\textstyle\sum_j z_j(\mathbf h)\mathbf d_j}_{\text{SAE feature components}}
  +\underbrace{\boldsymbol\epsilon(\mathbf h)}_{\text{reconstruction residual}}
  +\underbrace{\mathbf b_{\mathrm{dec}}}_{\text{decoder bias}},
  \label{eq:cos-sae-decomposition-main}
\end{equation}
where $z_j(\mathbf h)$ and $\mathbf d_j$ denote the activation and decoder direction of feature $j$, and $\boldsymbol\epsilon(\mathbf h)=\mathbf h-\hat{\mathbf h}$ is the SAE reconstruction residual. We quantify the contribution of each component to COS. Among the 50 most activated SAE features, we select the five with the largest COS contributions. Their maximally activating tokens are associated with reasoning transitions, intermediate conclusions, and final-answer expressions (Appendix~\ref{app:cos-sae}). We then intervene on each feature using feature clamping~\citep{templeton2024scaling}, setting it to zero at every generated-token position while preserving all other feature activations, decoder bias, and reconstruction residual.

Table~\ref{tab:sae-feature-ablation} shows that ablating all five selected features reduces accuracy on AIME24, with four interventions also lowering mean COS. Feature \texttt{31368} has the strongest effect, reducing accuracy from 30.00\% to 20.00\% and mean COS from 0.1238 to 0.1115.
The decomposition further shows that COS ranking is not explained by sparse features alone. Approximately 57\% of the ranking contribution comes from the reconstruction residual, 48\% from SAE features, and $-5\%$ from the decoder bias (Appendix~\ref{app:cos-sae}). Thus, COS combines interpretable reasoning- and answer-related features with substantial information not resolved by the current SAE.

\begin{table}[t]
\caption{SAE feature ablation on AIME24. \texttt{Numbers} denote SAE feature IDs.}

\label{tab:sae-feature-ablation}
\centering
\renewcommand{\arraystretch}{.75}
\setlength{\tabcolsep}{4mm}{
\scalebox{0.85}{
\begin{tabular}{@{}lcccc@{}}
\toprule
Condition & Accuracy (\%) & Mean COS & $\Delta$ Accuracy (pp) & $\Delta$ COS \\
\midrule
Baseline & 30.00  & 0.1238 & --- & --- \\
Ablate \texttt{31368} & 20.00  & 0.1115 & $-10.00$ & $-0.0123$ \\
Ablate \texttt{52642} & 26.67  & 0.1239 & $-3.33$ & $+0.0001$ \\
Ablate \texttt{41030} & 23.33 & 0.1212 & $-6.67$ & $-0.0026$ \\
Ablate \texttt{3447} & 23.33 & 0.1194 & $-6.67$ & $-0.0045$ \\
Ablate \texttt{48965} & 20.00  & 0.1220 & $-10.00$ & $-0.0018$ \\
\bottomrule
\end{tabular}}}

\end{table}

\begin{table}[t]
\caption{Comparison of direction-extraction datasets. Source Greedy is the model's accuracy on the source dataset. Pearson $r$ and Spearman $\rho$ between COS and question-level accuracy are averaged across MATH-500, AIME24, AIME25, and AMC.}

\label{tab:cos-source-comparison}
\centering
\renewcommand{\arraystretch}{.75}
\setlength{\tabcolsep}{5mm}{
\scalebox{0.8}{
\begin{tabular}{@{}lccc@{}}
\toprule
Direction source & Source Greedy (\%) & Mean Pearson $r$ & Mean Spearman $\rho$ \\
\midrule
PIQA & 87.65 & 0.6420 & 0.6273 \\
CommonsenseQA & 81.08 & 0.6340 & 0.6116 \\
MMLU-Pro & 72.69 & 0.6260 & 0.6071 \\
LogiQA 2.0 & 74.95 & 0.5942 & 0.5692 \\
GPQA & 44.95 & 0.5385 & 0.5126 \\
TruthfulQA & 71.97 & 0.5328 & 0.4519 \\
\bottomrule
\end{tabular}}}
\vspace{-0.5cm}
\end{table}

\subsection{Dependence on the Direction-Extraction Dataset (RQ4)}

\label{sec:cos-source}
To evaluate COS sensitivity to the source dataset, we extract directions from PIQA~\citep{bisk2020piqa}, CommonsenseQA~\citep{talmor2019commonsenseqa}, MMLU-Pro, LogiQA 2.0~\citep{liu2023logiqa2}, GPQA, and TruthfulQA using the same CAA procedure, and measure Pearson and Spearman correlations between COS and question-level accuracy on MATH-500, AIME24, AIME25, and AMC.
Table~\ref{tab:cos-source-comparison} shows positive correlations for all
six source datasets. PIQA and CommonsenseQA, on which the model achieves
87.65\% and 81.08\% source accuracy, yield the strongest average Spearman
correlations of 0.6273 and 0.6116, respectively. GPQA, with a lower source
accuracy of 44.95\%, yields a weaker average Spearman correlation of 0.5126.

We interpret this pattern as evidence that COS captures correctness-related internal representations. When the model reliably distinguishes correct from incorrect answers on the source task, the contrastive direction better aligns with objective correctness. On harder source tasks, teacher-forcing a correct answer does not guarantee that the model internally recognizes it as correct, weakening the extracted direction and its correlation with target-task accuracy.


These observations connect to the evidence-readout analysis in
Section~\ref{sec:SelectTTRL-theory}. When the model achieves higher source-task
accuracy, the activation contrast between correct and incorrect
answers is more likely to reflect differences in internal answer evidence.
The resulting CAA direction can read this evidence in
target tasks more effectively, improving COS ranking of question-level
accuracy. This interpretation motivates choosing a source dataset the model
solves with reasonably high accuracy.

\vspace{-0.3cm}
\section{Conclusion}

In this paper, we investigated how to improve the efficiency of Test-Time Reinforcement Learning by selecting test problems before training-time rollout collection and reinforcement learning. We identified COS, a representation-level signal that exhibited a stable relationship with problem-level empirical accuracy, and provided a theoretical analysis of this relationship. Based on this finding, we proposed SelectTTRL, which selected a subset of test problems for subsequent test-time training according to COS. Experiments across multiple reasoning benchmarks showed that SelectTTRL substantially reduced test-time computation while maintaining or improving the performance of standard TTRL. Overall, our results demonstrated the potential of internal model representations for more selective and efficient test-time adaptation.

\bibliography{iclr2027_conference,cttrl_method_refs}
\bibliographystyle{plainnat}

\newpage
\appendix
\section{Proofs and extensions of the COS analysis}
\label{app:cos-theory}

\subsection{Spherical model and response identity}
\label{app:cos-response}
Fix the model, layer, direction, prompt template, and reading protocol. Let $b_D$
be a probability density with respect to spherical surface measure
$\omega$, with support independent of $t$. For finite natural parameters,
define
\begin{equation}
  A_D(\eta)=\log\int_{\mathbb S^{d-1}}
  b_D(z)\exp(\eta^\top z)\,d\omega(z).
\end{equation}
Boundedness of $U$ permits differentiation under the integral. Hence
\begin{equation}
  \nabla A_D(\eta)=\mathbb E_\eta[U],\qquad
  \nabla^2 A_D(\eta)=\operatorname{Cov}_\eta(U).
\end{equation}
Applying the chain rule along $\eta=tu$ proves
Eq.~(\ref{eq:cos-response}). More generally, for any bounded measurable~$f$,
\begin{equation}
  \frac{d}{dt}\mathbb E_t[f(U)]
  =\operatorname{Cov}_t(f(U),u^\top U).
  \label{eq:cos-general-response}
\end{equation}
These are standard exponential-family identities
\citep{wainwright2008graphical}.

The model also arises by minimizing $D_{\rm KL}(p\|b_D)$ subject to
normalization and a specified expectation of $u^\top U$. A Lagrange
multiplier for the latter constraint yields the exponential tilt in
Eq.~(\ref{eq:cos-evidence-model}).
With uniform $b_D$, the distribution is vMF with concentration $|t|$
and direction $\operatorname{sign}(t)u$ for $t\ne0$. If
$b_D(U)\propto\exp(a_D^\top U)$, its vMF natural parameter is $a_D+tu$,
so both direction and concentration may vary.

\subsection{From raw source contrasts to a spherical response}
\label{app:cos-source}
For the empirical raw mean difference $\hat d_S\ne0$, Cauchy--Schwarz gives
\begin{equation}
  \frac{\hat d_S}{\|\hat d_S\|_2}
  =\arg\max_{\|w\|_2=1}
  \frac1n\sum_{i=1}^n w^\top(h_i^+-h_i^-).
\end{equation}
Thus the extracted direction maximizes the mean signed raw-activation contrast.

For its evidence interpretation, consider the source construction
$h^\pm=c+R^\pm V^\pm+\zeta^\pm$, where $c$ is a paired common component,
$R^\pm\ge0$ have finite means, and $V^\pm$ are unit vectors following
the source spherical family at $t_+>t_-$. Here $V^\pm$ describe the
varying source component.
Set $r_\pm=\mathbb E[R^\pm]$, $m_\pm=\mathbb E[V^\pm]$, and
$\bar R_S=(r_++r_-)/2>0$. Define the vector covariances
$k_\pm=\mathbb E[(R^\pm-r_\pm)(V^\pm-m_\pm)]$. Then
\begin{equation}
\begin{aligned}
  d_S^\star
  &=\mathbb E[h^+-h^-]=\bar R_S(m_+-m_-)+e_S,\\
  e_S
  &=(r_+-\bar R_S)m_+-(r_--\bar R_S)m_-
    +k_+-k_-+\mathbb E[\zeta^+-\zeta^-].
\end{aligned}
\end{equation}
Integrating Eq.~(\ref{eq:cos-response}) in the source domain proves
Eq.~(\ref{eq:cos-source-response}). Each $\Sigma_S(r)$ is positive
semidefinite. Thus $u^\top M_Su>0$ when
$\operatorname{Var}_r(u^\top V)>0$ on a set of positive integration measure.
If $e_S=0$, this gives $u^\top d_S^\star>0$. Axis symmetry of $b_S$
about $u$ further gives $d_S^\star=\gamma_Su$ with $\gamma_S>0$.

Equal radial means, radial independence, and zero mean unmatched contrast
give $e_S=0$. More generally, the transfer analysis incorporates nonzero
source bias and finite-sample error through the actual extracted direction.

\subsection{Structural hypotheses and source reliability}
\label{app:cos-hypotheses}

\paragraph{Uniform source gain.}
Under $\mathsf H_1$, for $d_0+\xi_S\ne0$,
\begin{equation}
  v_S=\frac{d_0+\xi_S}{\|d_0+\xi_S\|_2}.
\end{equation}
Since $\xi_S$ is independent of source accuracy $a_S$, so is $v_S$.
On a fixed target representation cache, any readout-quality statistic
$\mathcal R_D(v_S)$ computed with fixed correctness labels is therefore
independent of $a_S$, whenever defined. This includes Pearson and Spearman
correlations. The overall trend in Table~\ref{tab:cos-source-comparison}
argues against this independence prediction.

\paragraph{Single-factor linear response.}
One realization of $\mathsf H_2$ assumes a common intervention direction $w$,
$\mathbb E_j[U]-\mathbb E_0[U]=\tau_jw$, and a shared affine reliability
function $P_j=p_0+\eta\mathbb E_j[T]$, with $\eta>0$ and values in $[0,1]$.
Then
\begin{equation}
  \Delta C_j=\tau_jv^\top w,\qquad
  \Delta P_j=\eta\tau_ju^\top w.
\end{equation}
When $v^\top w$ and $u^\top w$ have the same nonzero sign, eliminating
$\tau_j$ gives Eq.~(\ref{eq:cos-single-factor}) with
$b=\eta(u^\top w)/(v^\top w)>0$. On the natural evidence path, the analogous
closure $\Sigma_D(t)u=\lambda(t)w$, $\lambda(t)>0$, gives a positive COS
response whenever $v^\top w>0$.
Table~\ref{tab:sae-feature-ablation} departs from the predicted proportionality:
ablations 31368 and 48965 both reduce accuracy by 10 percentage points,
whereas their mean COS reductions are 0.0123 and 0.0018.

\paragraph{Positive compatibility and proof of Lemma~\ref{lem:cos-positive-response}.}
The orthogonal decomposition of the realized contrast is exact:
$\alpha_S=u^\top\hat d_S$ and $r_S=(I-uu^\top)\hat d_S$.
Consequently,
\begin{equation}
  v^\top\Sigma_D(t)u
  =\frac{\alpha_S\beta_D(t)+r_S^\top\Sigma_D(t)u}
         {\|\hat d_S\|_2}
  \geq\frac{\delta\alpha_S\beta_D(t)}{\|\hat d_S\|_2}>0.
\end{equation}
The inequality is precisely Condition~\ref{cond:cos-compatibility}.
Together with Eq.~(\ref{eq:cos-response}), it proves the lemma.
If $\Sigma_D(t)=\sum_k\lambda_k w_kw_k^\top$, the numerator before
normalization is $\sum_k\lambda_k(\hat d_S^\top w_k)(u^\top w_k)$.
Here $w_k$ are covariance eigenvectors, and compatibility constrains
their aggregate contribution.

\paragraph{A finite-sample sufficient condition.}
Write the population contrast in Eq.~(\ref{eq:cos-source-response}) as
$d_S^\star=s_Su+r_S^\star$, with $r_S^\star\perp u$ and $s_S>0$.
In terms of the source geometry,
$s_S=u^\top M_Su+u^\top e_S$ and
$r_S^\star=(I-uu^\top)(M_Su+e_S)$.
Let $\hat d_S=d_S^\star+\epsilon_S$. For $n$ i.i.d. source contrasts
with mean $d_S^\star$ and finite covariance $\Omega_S$,
\begin{equation}
  \mathbb E[\epsilon_S]=0,\qquad
  \mathbb E\|\epsilon_S\|_2^2=\frac{\operatorname{tr}\Omega_S}{n}.
\end{equation}
Define the target anisotropy ratio
\begin{equation}
  \chi_*=\sup_t
  \frac{\|(I-uu^\top)\Sigma_D(t)u\|_2}{\beta_D(t)}.
\end{equation}
For any $0<\tau<s_S$, Markov's inequality gives
$P(\|\epsilon_S\|_2\leq\tau)\geq
1-\operatorname{tr}(\Omega_S)/(n\tau^2)$.
On this event, $\alpha_S\geq s_S-\tau$ and
$\|r_S\|_2\leq\|r_S^\star\|_2+\tau$. Thus the sufficient condition
\begin{equation}
  \chi_*\bigl(\|r_S^\star\|_2+\tau\bigr)
  \leq(1-\delta)(s_S-\tau)
  \label{eq:cos-source-quality-margin}
\end{equation}
implies positive compatibility with the same probability bound, since
$r_S^\top\Sigma_D(t)u\geq
-\|r_S\|_2\chi_*\beta_D(t)$.
Holding residual bias, sample size, and noise fixed, stronger evidence
$s_S$ increases this margin. This provides a mechanism by which source
reliability can improve transfer through smaller relative extraction error.
Source accuracy provides an empirical proxy for source reliability;
the margin also depends on residual bias and extraction noise.

\subsection{Evidence retained by a projection}
\begin{coslemma}[Evidence retained by a projection]
\label{lem:cos-information}
Under Assumption~\ref{ass:cos-evidence}, $T=u^\top U$ is sufficient for
the evidence-state parameter $t$. Let $J_U(t)$ and $J_X(t)$ denote the
Fisher information about $t$ in $U$ and $X$, respectively. Whenever the
denominator is positive,
\begin{equation}
  \frac{J_X(t)}{J_U(t)}
  \geq
  \frac{\bigl(v^\top\Sigma_D(t)u\bigr)^2}
       {\bigl(v^\top\Sigma_D(t)v\bigr)
        \bigl(u^\top\Sigma_D(t)u\bigr)}.
  \label{eq:cos-information-retention}
\end{equation}
\end{coslemma}

\paragraph{Proof.}
The density factorizes through $T=u^\top U$, so $T$ is sufficient for $t$
within the model. The likelihood score for the full representation is
\begin{equation}
  \partial_t\log p_D(U\mid t)=T-\mathbb E_t[T],
  \qquad J_U(t)=\operatorname{Var}_t(T).
\end{equation}
For the induced distribution of $X=v^\top U$, differentiation of the
marginal gives the score $\mathbb E_t[T-\mathbb E_tT\mid X]$. Therefore
\begin{equation}
  J_X(t)=\operatorname{Var}_t\bigl(\mathbb E_t[T\mid X]\bigr).
\end{equation}
Conditional expectation and Cauchy--Schwarz imply
\begin{equation}
\begin{aligned}
  \operatorname{Cov}_t(X,T)^2
  &=\operatorname{Cov}_t\bigl(X,\mathbb E_t[T\mid X]\bigr)^2\\
  &\leq \operatorname{Var}_t(X)J_X(t).
\end{aligned}
\end{equation}
Substituting the projected covariances proves
Eq.~(\ref{eq:cos-information-retention}), quantifying information retained
about the evidence-state parameter $t$. Condition~\ref{cond:cos-compatibility}
additionally establishes the positive response direction.

\subsection{Proof of Theorem~\ref{thm:cos-transfer}}
Under the reliability condition, $p_q=G_D(t_q)$, while
$\mu_q=\mathbb E[X\mid q]=F_D(t_q)$. Lemma~\ref{lem:cos-positive-response} gives
$F_D'(t)>0$. For independent $T,T'$ from the same state,
\begin{equation}
  G_D'(t)=\operatorname{Cov}_t(g_D(T),T)
  =\tfrac12\mathbb E_t[(g_D(T)-g_D(T'))(T-T')]>0.
\end{equation}
Strict positivity follows from strict increase of $g_D$ and nonzero
variance of $T$. For independent question states $t_Q,t_{Q'}$,
\begin{equation}
\begin{aligned}
  C_0&:=\operatorname{Cov}_Q(F_D(t_Q),G_D(t_Q))\\
  &=\tfrac12\mathbb E\bigl[
  (F_D(t_Q)-F_D(t_{Q'}))(G_D(t_Q)-G_D(t_{Q'}))\bigr]>0.
\end{aligned}
\end{equation}
Each scoring response has conditional mean $\mu_q$, so
$\mathbb E[c_K\mid Q]=\mu_Q$. Conditional independence of $Y'$ and the
scoring responses then gives
\begin{equation}
  \operatorname{Cov}(c_K,Y')
  =\operatorname{Cov}_Q(\mu_Q,p_Q)=C_0
  =\operatorname{Cov}(c_K,p_Q).
\end{equation}
Nonzero covariance implies dependence, hence $I(c_K;Y')>0$.
All variables in this covariance calculation are bounded.

\subsection{Angular tolerance and departures from the model}
\label{app:cos-robustness}
\paragraph{Angular tolerance.}
Write $v=\cos\theta\,u+\sin\theta\,w$ with $w\perp u$ and $\|w\|_2=1$.
For $0\leq\theta<\pi/2$, define
\begin{equation}
  \chi_D(t)=
  \frac{\|(I-uu^\top)\Sigma_D(t)u\|_2}
       {u^\top\Sigma_D(t)u}.
\end{equation}
Projection and Cauchy--Schwarz yield
\begin{equation}
  F_D'(t)\geq
  \bigl(\cos\theta-\chi_D(t)\sin\theta\bigr)
  u^\top\Sigma_D(t)u.
\end{equation}
Thus $\cos\theta>\sup_t\chi_D(t)\sin\theta$ suffices for positive
transfer, allowing angular mismatch under anisotropic target geometry.

\paragraph{Question heterogeneity.}
Let actual question means and success probabilities deviate from the
ideal family as
$\mu_q=F_D(t_q)+e_q$ and $p_q=G_D(t_q)+r_q$.
With $\delta_x=\operatorname{sd}(e_Q)$,
$\delta_p=\operatorname{sd}(r_Q)$,
$\sigma_F=\operatorname{sd}(F_D(t_Q))$, and
$\sigma_G=\operatorname{sd}(G_D(t_Q))$, expansion gives
\begin{equation}
\begin{aligned}
  \operatorname{Cov}_Q(\mu_Q,p_Q)
  &\geq C_0-\delta_x\sigma_G-\delta_p\sigma_F-\delta_x\delta_p.
  \label{eq:cos-perturbation}
\end{aligned}
\end{equation}
Positive association survives whenever the right-hand side is positive,
even if individual question pairs reverse order. This lower bound allows
arbitrary signs of the heterogeneity cross terms.

\paragraph{Averaging.}
For conditionally independent, identically distributed scoring responses,\newline
write $C=\operatorname{Cov}_Q(\mu_Q,p_Q)$ and
$\sigma_q^2=\operatorname{Var}(X\mid q)$. Then
\begin{equation}
  \operatorname{Corr}(c_K,p_Q)
  =\frac{C}{\sqrt{[\operatorname{Var}_Q(\mu_Q)
      +\mathbb E_Q\sigma_Q^2/K]\operatorname{Var}_Q(p_Q)}}.
\end{equation}
For $C>0$, averaging increases this population correlation by reducing
sampling variance.

\paragraph{Greedy scores.}
\label{app:cos-greedy}
For the greedy score, define
$d_q=c_{\rm g}(q)-\mu_q$,
\begin{equation}
  \operatorname{Cov}_Q(c_{\rm g}(Q),p_Q)
  =C+\operatorname{Cov}_Q(d_Q,p_Q).
\end{equation}
Its association remains positive if $\operatorname{Cov}_Q(d_Q,p_Q)>-C$.

\subsection{Empirical support for question-level association}
We examine the positive association in Theorem~\ref{thm:cos-transfer}
and its persistence under question heterogeneity with Qwen3-8B on
MATH-500 (500 questions), GPQA-Diamond (198), MMLU-Pro STEM (498),
and AIME 2024 (30). We use the layer-21 TruthfulQA direction and
16 sampled responses per question. COS is read at the closing token
of the final complete boxed answer, or the response endpoint when no
complete box is available. Splitting responses by sample index into
two groups of eight gives $(c_A,\widehat p_A)$ and $(c_B,\widehat p_B)$.
Cross-half estimates use disjoint responses for COS and accuracy,
matching the theorem's conditional-independence construction.

\paragraph{Projection gap and cross-response association.}
The class-projection gap $\beta=\mathbb E[X\mid Y=1]-\mathbb E[X\mid Y=0]$
is positive on all four datasets, with 95\% confidence intervals above
zero. Both cross-half Spearman correlations are also positive
(Table~\ref{tab:cos-projection-diagnostics}).

\begin{table}[t]
\centering
\caption{Class-projection gaps and cross-half COS--accuracy correlations.
Confidence intervals for $\widehat\beta$ use 1,000 question-level
bootstrap resamples, retaining all responses of each sampled question.}
\label{tab:cos-projection-diagnostics}
\small
\setlength{\tabcolsep}{5pt}
\begin{tabular}{@{}lrrrr@{}}
\toprule
Dataset & $\widehat\beta$ & 95\% CI
& $\rho(c_A,\widehat p_B)$ & $\rho(c_B,\widehat p_A)$ \\
\midrule
MATH-500 & 0.02271 & $[0.01815,\,0.02821]$ & 0.296 & 0.338 \\
GPQA-Diamond & 0.01602 & $[0.01125,\,0.02081]$ & 0.249 & 0.244 \\
MMLU-Pro STEM & 0.02131 & $[0.01803,\,0.02460]$ & 0.434 & 0.464 \\
AIME 2024 & 0.05077 & $[0.03646,\,0.06377]$ & 0.536 & 0.610 \\
\bottomrule
\end{tabular}
\end{table}

\paragraph{Association under question heterogeneity.}
With $\alpha=\mathbb E[X\mid Y=0]$, define
$\varepsilon_q=\mu_q-\alpha-\beta p_q$. Then
\begin{equation}
  C=\operatorname{Cov}_Q(\mu_Q,p_Q)
   =\beta\operatorname{Var}_Q(p_Q)
    +\operatorname{Cov}_Q(\varepsilon_Q,p_Q).
  \label{eq:SelectTTRL-score-decomposition}
\end{equation}
We estimate $\alpha,\beta$ from class-conditional projection means using
five-fold cross-fitting over questions. Within each held-out fold,
\begin{equation}
\begin{aligned}
  \widehat C_{\mathrm{total}}
  &=\tfrac12\bigl[\widehat{\operatorname{Cov}}(c_A,\widehat p_B)
    +\widehat{\operatorname{Cov}}(c_B,\widehat p_A)\bigr],\\
  \widehat C_{\mathrm{proj}}
  &=\widehat\beta\,\widehat{\operatorname{Cov}}(\widehat p_A,\widehat p_B),
  \qquad \widehat C_{\mathrm{res}}
  =\widehat C_{\mathrm{total}}-\widehat C_{\mathrm{proj}}.
\end{aligned}
\label{eq:cos-cross-half-decomposition}
\end{equation}
Sample covariances are averaged with fold-size weights. Confidence intervals
use 1,000 within-fold question bootstrap resamples, refitting
$\widehat\alpha,\widehat\beta$ each time.

\begin{table}[t]
\centering
\caption{Cross-half covariance decomposition. All values are multiplied
by $10^3$; brackets give 95\% bootstrap confidence intervals.}
\label{tab:cos-covariance-diagnostics}
\small
\setlength{\tabcolsep}{7pt}
\renewcommand{\arraystretch}{1.35}
\begin{tabular}{@{}lccc@{}}
\toprule
Dataset & $\widehat C_{\mathrm{total}}$
& $\widehat C_{\mathrm{proj}}$ & $\widehat C_{\mathrm{res}}$ \\
\midrule
MATH-500 & \shortstack{2.158\\$[1.535,\,2.834]$}
& \shortstack{1.752\\$[1.276,\,2.267]$}
& \shortstack{0.407\\$[0.136,\,0.673]$} \\
GPQA-Diamond & \shortstack{1.884\\$[0.993,\,2.842]$}
& \shortstack{1.879\\$[1.302,\,2.425]$}
& \shortstack{0.005\\$[-0.591,\,0.627]$} \\
MMLU-Pro STEM & \shortstack{2.909\\$[2.349,\,3.491]$}
& \shortstack{2.471\\$[2.017,\,2.927]$}
& \shortstack{0.438\\$[0.169,\,0.708]$} \\
AIME 2024 & \shortstack{7.533\\$[2.509,\,9.894]$}
& \shortstack{6.738\\$[2.290,\,8.874]$}
& \shortstack{0.796\\$[-1.364,\,2.678]$} \\
\bottomrule
\end{tabular}
\end{table}

The total covariance is positive on all four datasets, with confidence
intervals above zero, and the projection term accounts for most of it
(Table~\ref{tab:cos-covariance-diagnostics}). The heterogeneity term is
positive on MATH-500 and MMLU-Pro STEM; its intervals include zero on
GPQA-Diamond and AIME 2024. Thus positive question-level association
persists alongside heterogeneous response representations.

\section{SAE Decomposition and Feature Interventions for COS}
\label{app:cos-sae}

\subsection{Exact decomposition}
Let $\mathbf h$ be the nonzero residual-stream representation used to
compute COS, and let $z_j(\mathbf h)$ and $\mathbf d_j$ denote the
activation and decoder direction of SAE feature $j$. Following
\citet{marks2025sparsecircuits}, write
\begin{equation}
  \mathbf h=\sum_j z_j(\mathbf h)\mathbf d_j
  +\mathbf b_{\mathrm{dec}}+\boldsymbol\epsilon(\mathbf h),
  \qquad
  \boldsymbol\epsilon(\mathbf h)=\mathbf h-\operatorname{SAE}(\mathbf h).
  \label{eq:cos-sae-reconstruction}
\end{equation}
All terms are expressed in the same activation coordinates. For the
unit correctness-related direction $\mathbf v_\ell$, define
\begin{equation}
\begin{aligned}
  c_j(\mathbf h)&=\frac{z_j(\mathbf h)\mathbf d_j^\top\mathbf v_\ell}
  {\|\mathbf h\|_2},\\
  c_{\mathrm{bias}}(\mathbf h)&=\frac{\mathbf b_{\mathrm{dec}}^\top\mathbf v_\ell}
  {\|\mathbf h\|_2},\qquad
  c_{\mathrm{res}}(\mathbf h)=\frac{\boldsymbol\epsilon(\mathbf h)^\top\mathbf v_\ell}
  {\|\mathbf h\|_2}.
\end{aligned}
\label{eq:cos-sae-contributions}
\end{equation}
Then $\mathrm{COS}(\mathbf h)=\sum_j c_j(\mathbf h)
+c_{\mathrm{bias}}(\mathbf h)+c_{\mathrm{res}}(\mathbf h)$ exactly.
The residual uses the full SAE reconstruction, and all contributions
share the norm of the original representation as their denominator.

\subsection{Ranking contribution accounting}
We quantify ranking contributions through the COS difference between
high- and low-scoring questions on AIME24. After excluding truncated
greedy responses, we rank questions by answer-position COS and select
the five highest and five lowest as groups $H$ and $L$, respectively.
For each question, let $c_{\mathrm{SAE}}=\sum_j c_j$, with
$c_{\mathrm{bias}}$ and $c_{\mathrm{res}}$ defined in
Eq.~(\ref{eq:cos-sae-contributions}). For each component
$k\in\{\mathrm{SAE},\mathrm{bias},\mathrm{res}\}$, define
\begin{equation}
\begin{aligned}
  \Delta_k
  &=\frac{1}{5}\sum_{q\in H}c_k(\mathbf h_q)
    -\frac{1}{5}\sum_{q\in L}c_k(\mathbf h_q),\\
  w_k&=\frac{\Delta_k}{\Delta_{\mathrm{COS}}},\qquad
  \Delta_{\mathrm{COS}}=\sum_k\Delta_k.
\end{aligned}
\label{eq:cos-ranking-contribution}
\end{equation}
Here $\mathbf h_q$ is the answer-position representation of the greedy
response. The signed shares sum to one; a negative share indicates
that the component offsets the high--low COS difference.

\begin{table}[t]
\centering
\caption{Component contributions to the high--low COS difference on
AIME24. Each group contains five questions. Values are rounded.}
\label{tab:cos-ranking-decomposition}
\small
\setlength{\tabcolsep}{6pt}
\begin{tabular}{@{}lrrrr@{}}
\toprule
Component & High COS & Low COS & Difference & Share (\%) \\
\midrule
SAE features & 0.07968 & 0.05503 & $+0.02465$ & 47.6 \\
Decoder bias & 0.05289 & 0.05538 & $-0.00249$ & $-4.8$ \\
Reconstruction residual & 0.04675 & 0.01714 & $+0.02962$ & 57.2 \\
\midrule
Total COS & 0.17933 & 0.12755 & $+0.05178$ & 100.0 \\
\bottomrule
\end{tabular}
\end{table}

As shown in Table~\ref{tab:cos-ranking-decomposition}, SAE features
account for 47.6\% of the COS difference and the reconstruction residual
for 57.2\%, while the decoder bias offsets 4.8\%. Thus, both sparse
features and information retained in the reconstruction residual
contribute to the observed COS ranking.

\subsection{Feature clamping}
We select the five features with the largest absolute differences in
mean COS contribution between the five highest- and five lowest-COS questions.
Following \citet{templeton2024scaling}, at each generated-token position
$t$ we replace the activation of a selected feature $j$ by zero while
retaining the other local SAE activations, decoder bias, and original
reconstruction residual. The modified activation is
\begin{equation}
  \mathbf h'_t
  =\sum_{k\ne j}z_k(\mathbf h_t)\mathbf d_k
    +\mathbf b_{\mathrm{dec}}+\boldsymbol\epsilon(\mathbf h_t)
  =\mathbf h_t-z_j(\mathbf h_t)\mathbf d_j.
\label{eq:cos-sae-clamping}
\end{equation}
The model continues its forward pass and autoregressive generation using
this modified activation.

\subsection{Feature activation examples}
\label{app:cos-feature-examples}
Given the small number of AIME24 questions, we use the larger MATH-500
set to illustrate token-wise SAE activations in original responses.
Table~\ref{tab:cos-feature-examples} presents representative high-activation
locations, including goal expressions, result statements, numerical
outputs, and formula boundaries.

\begin{table}[H]
\centering
\caption{Examples of high-activation locations for COS-related SAE
features in MATH-500 responses.}
\label{tab:cos-feature-examples}
\small
\setlength{\tabcolsep}{5pt}
\renewcommand{\arraystretch}{1.3}
\begin{tabular}{@{}lp{0.43\linewidth}p{0.29\linewidth}@{}}
\toprule
Feature ID & Observed text or symbol & Location \\
\midrule
\texttt{32531}, \texttt{47984} & \texttt{\$\$} following
\texttt{\textbackslash boxed\{6\}} or
\texttt{\textbackslash boxed\{3\}}
& Formula-ending delimiter after the final answer \\
\texttt{3447} & \texttt{So}; \texttt{This}; \texttt{exact expression is}
& Connective or result-statement expression \\
\texttt{41030} & \texttt{58500}; \texttt{2000}; \texttt{4 inches per second}
& Numerical results and units \\
\texttt{31368} & \texttt{find the sum}; \texttt{how many};
\texttt{ordered quadruple}
& Expression of the quantity or object to be found \\
\texttt{605} & The digit \texttt{7} in \texttt{\textbackslash boxed\{27\}}
& Numerical token within the final answer \\
\bottomrule
\end{tabular}
\end{table}

\FloatBarrier
\section{Training Efficiency}
\label{app:training-efficiency}

Table~\ref{tab:training-efficiency} reports the training-cost reductions
of SelectTTRL relative to TTRL with Qwen3-8B across six benchmarks.

\begin{table}[H]
\caption{Training-efficiency reductions and time savings across six benchmarks.}
\label{tab:training-efficiency}
\centering
\small
\setlength{\tabcolsep}{4pt}
\renewcommand{\arraystretch}{1.}
\begin{tabular}{@{}lrrrrr@{}}
\toprule
& \multicolumn{3}{c}{Reduction (\%)} & \multicolumn{2}{c}{Time saved} \\
\cmidrule(lr){2-4}\cmidrule(l){5-6}
Dataset & \shortstack{Generated\\tokens}
& \shortstack{Update response\\tokens}
& \shortstack{Training-step\\time}
& \shortstack{Wall-clock\\(min)} & GPU-hours \\
\midrule
MATH-500 & 17.07 & 17.14 & 5.43 & 14.23 & 0.949 \\
MMLU-Pro STEM & 19.56 & 18.93 & 6.68 & 18.02 & 1.202 \\
GPQA-Diamond & 2.88 & 3.24 & 0.66 & 1.43 & 0.095 \\
AIME 2024 & 10.74 & 10.73 & 3.73 & 12.96 & 0.864 \\
AIME 2025 & 26.12 & 25.93 & 10.86 & 36.66 & 2.444 \\
AMC & 27.83 & 27.91 & 8.91 & 35.47 & 2.365 \\
\midrule
\textbf{Aggregate} & \textbf{18.38} & \textbf{18.33} & \textbf{6.48}
& \textbf{118.77} & \textbf{7.918} \\
\bottomrule
\end{tabular}
\end{table}

\end{document}